\documentclass[dvipsnames]{article}

\usepackage[final]{openbmb_2025}

\usepackage{amsmath}
\usepackage{amssymb}
\usepackage{mathtools}
\usepackage{amsthm}
\usepackage{multirow}
\usepackage{makecell}
\usepackage{caption}
\usepackage{subcaption}
\usepackage{wrapfig2}
\usepackage{graphicx}
\usepackage{pifont}
\usepackage{siunitx}

\usepackage[table]{xcolor}
\usepackage{tabularx}
\usepackage{array}
\usepackage{siunitx}
\usepackage{fvextra}
\usepackage{stmaryrd}

\usepackage[most]{tcolorbox}
\newtcolorbox[auto counter, number within=section]{examplebox}[2][]{%
    enhanced,
    breakable,
    fonttitle=\bfseries,
    title=#2,
    label={#1},
    colback=white,
    colframe=black!75,          
    colbacktitle=black!65,      
    coltitle=white
}

\definecolor{leanKeyword}{HTML}{1F4E79}   
\definecolor{leanCommand}{HTML}{7A3E9D}   
\definecolor{leanTactic}{HTML}{0F766E}    
\definecolor{leanComment}{HTML}{6A737D}   
\definecolor{leanString}{HTML}{A05A2C}    
\definecolor{leanSymbol}{HTML}{374151}    
\lstdefinelanguage{lean4}{
    sensitive=true,
    alsoletter={_'},
    morecomment=[l]{--},
    morecomment=[s]{/-}{-/},
    morestring=[b]",
    morekeywords=[1]{
        import,open,namespace,section,end,variable,variables,universe,universes,
        theorem,lemma,example,def,abbrev,axiom,opaque,inductive,coinductive,
        structure,class,instance,where,deriving,syntax,macro,notation,
        scoped,local,mutual,partial,noncomputable
    },
    morekeywords=[2]{
        by,fun,let,have,show,suffices,from,if,then,else,match,with,
        do,for,in,return,try,catch
    },
    morekeywords=[3]{
        intro,intros,apply,exact,refine,rw,simp,dsimp,constructor,
        cases,induction,subst,rfl,assumption,ring,omega,linarith,
        norm_num,aesop,unfold,decide,contradiction,trivial,left,right,
        use,obtain,rcases
    },
    keywordstyle=[1]\color{leanCommand}\bfseries,
    keywordstyle=[2]\color{leanKeyword}\bfseries,
    keywordstyle=[3]\color{leanTactic}\bfseries,
    commentstyle=\color{leanComment}\itshape,
    stringstyle=\color{leanString},
    basicstyle=\ttfamily\small\color{leanSymbol},
    breaklines=true,
    breakatwhitespace=false,
    showstringspaces=false,
    columns=flexible,
    keepspaces=true,
    extendedchars=true,
    literate=
        {ℝ}{{$\mathbb{R}$}}1
        {ℂ}{{$\mathbb{C}$}}1
        {ℕ}{{$\mathbb{N}$}}1
        {ℤ}{{$\mathbb{Z}$}}1
        {ℚ}{{$\mathbb{Q}$}}1    
        {π}{{$\pi$}}1
        {��}{{$\mathcal{N}$}}1
        {ᶜ}{{$^{\mathrm{c}}$}}1
        {₀}{{$_0$}}1
        {₁}{{$_1$}}1
        {₂}{{$_2$}}1
        {₃}{{$_3$}}1
        {₄}{{$_4$}}1
        {₅}{{$_5$}}1
        {₆}{{$_6$}}1
        {₇}{{$_7$}}1
        {₈}{{$_8$}}1
        {₉}{{$_9$}}1
        {∈}{{$\in$}}1
        {∉}{{$\notin$}}1
        {⊆}{{$\subseteq$}}1
        {⊂}{{$\subset$}}1
        {∃}{{$\exists$}}1
        {∀}{{$\forall$}}1
        {≤}{{$\le$}}1
        {≥}{{$\ge$}}1
        {≠}{{$\ne$}}1
        {∧}{{$\wedge$}}1
        {∨}{{$\vee$}}1
        {¬}{{$\neg$}}1
        {→}{{$\to$}}1
        {↔}{{$\leftrightarrow$}}1
        {λ}{{$\lambda$}}1
        {·}{{$\cdot$}}1
        {⟨}{{$\langle$}}1
        {⟩}{{$\rangle$}}1
        {‖}{{$\Vert$}}1
        {ν}{{$\nu$}}1
        {α}{{$\alpha$}}1
        {β}{{$\beta$}}1
        {γ}{{$\gamma$}}1
        {δ}{{$\delta$}}1
        {ε}{{$\varepsilon$}}1
        {θ}{{$\theta$}}1
        {λ}{{$\lambda$}}1
        {μ}{{$\mu$}}1
        {ν}{{$\nu$}}1
        {π}{{$\pi$}}1
        {ρ}{{$\rho$}}1
        {σ}{{$\sigma$}}1
        {τ}{{$\tau$}}1
        {φ}{{$\varphi$}}1
        {ψ}{{$\psi$}}1
        {ω}{{$\omega$}}1
        {×}{{$\times$}}1
        {⊤}{{$\top$}}1
        {⊥}{{$\bot$}}1
        {•}{{$\cdot$}}1
        {∑}{{$\textstyle\sum$}}1
        {∏}{{$\prod$}}1
        {∫}{{$\int$}}1
        {ˢ}{{$^{\mathrm{s}}$}}1
        {∂}{{$\partial$}}1
        {⨅}{{$\bigsqcap$}}1
        {⨆}{{$\bigsqcup$}}1
        {⊓}{{$\sqcap$}}1
        {⊔}{{$\sqcup$}}1
        {◁}{{$\triangleleft$}}1
        {▷}{{$\triangleright$}}1
        {≃}{{$\simeq$}}1
        {≅}{{$\cong$}}1
        {⁅}{{$[\![$}}2
        {⁆}{{$]\!]$}}2
        {⁻¹}{{$^{-1}$}}2
}

\usepackage{newunicodechar}

\usepackage[utf8]{inputenc} 
\usepackage[T1]{fontenc}    
\usepackage{url}            
\usepackage{booktabs}       
\usepackage{amsfonts}       
\usepackage{nicefrac}       
\usepackage{microtype}      

\usepackage{color, soul}
\usepackage[most,breakable]{tcolorbox}
\usepackage{xcolor}         
\usepackage{listings}
\usepackage{amssymb} 

\usepackage[labelfont=bf]{caption}
\usepackage{enumitem}
\usepackage{tablefootnote}
\usepackage{threeparttable}
\usepackage{tabularx}
\usepackage{booktabs}
\usepackage{colortbl} 
\usepackage{placeins}
\definecolor{lightgray}{gray}{0.95}
\definecolor{oursrow}{HTML}{DCEEFA} 

\usepackage{fancyhdr}
\renewcommand{\headrulewidth}{1pt}
\makeatletter
\def\headrule{{\if@fancyplain\let\headrulewidth\plainheadrulewidth\fi
\hrule\@height\headrulewidth\@width\textwidth \vskip-\headrulewidth}}
\makeatother

\definecolor{BMBDarkBlue}{HTML}{315EFE}
\definecolor{BMBLightBlue}{HTML}{00D3ED}

\usepackage[colorlinks, linkcolor=RoyalBlue, anchorcolor=RoyalBlue, citecolor=RoyalBlue, urlcolor=RoyalBlue]{hyperref}
\usepackage{amsthm}

\renewcommand{\hflink}{https://huggingface.co/datasets/openbmb/FormalVerse}
\newcommand{\hflinkmodel}{https://huggingface.co/openbmb/MathForm-8B}
\renewcommand{\githublink}{https://github.com/openbmb/MathForm}
\makeatletter
\renewcommand{\@maketitle}{%
  \vbox{%
    \hsize\textwidth
    \linewidth\hsize
    \vskip 0.1in
    \begin{center}
    {\LARGE\bf \@title\par}
    \vspace{-1em}
    \if@submission
      \begin{tabular}[t]{c}\bf\rule{\z@}{24\p@}
        Anonymous Author(s) \\[6pt]
        Affiliation \\
        Address \\
        \texttt{email} \\
      \end{tabular}%
    \else
      \def\And{%
        \end{tabular}\hfil\linebreak[0]\hfil%
        \begin{tabular}[t]{c}\bf\rule{\z@}{24\p@}\ignorespaces%
      }
      \def\AND{%
        \end{tabular}\hfil\linebreak[4]\hfil%
        \begin{tabular}[t]{c}\bf\rule{\z@}{24\p@}\ignorespaces%
      }
      \begin{tabular}[t]{c}\bf\rule{\z@}{24\p@}\@author\end{tabular}%
    \fi
    \vskip 0.3in \@minus 0.1in
  \vspace{-2.0em}
  \begin{tabular}{rl}
    \huggingface & \url{\hflink} \\
    & \url{\hflinkmodel} \\
    \github & \url{\githublink}
  \end{tabular}
  \end{center}
  }
}
\makeatother

\newtcolorbox{mytheorem}{
  colback=gray!5, 
  colframe=gray!80, 
  boxrule=0.5pt, 
  arc=4pt, 
  left=4pt, 
  right=4pt, 
  top=4pt, 
  bottom=4pt, 
}

\lstdefinestyle{prompt}{%
    basicstyle={\footnotesize\ttfamily},
    numbers=none,
    xleftmargin=0pt,
    framexleftmargin=0pt,
    breakindent=0pt,
    columns=fullflexible,
    keepspaces=true,
    showstringspaces=false,
    tabsize=2,
    breaklines=true,
    breakatwhitespace=false,
    mathescape=true,
    literate={`}{{\char`\`}}1 {'}{{\char39}}1
             {ℝ}{{$\mathbb{R}$}}1 {ℕ}{{$\mathbb{N}$}}1
             {ℤ}{{$\mathbb{Z}$}}1 {ℚ}{{$\mathbb{Q}$}}1
             {≤}{{$\leq$}}1 {⊆}{{$\subseteq$}}1 {⊓}{{$\sqcap$}}1
             {⊥}{{$\bot$}}1 {∀}{{$\forall$}}1 {∃}{{$\exists$}}1
             {∈}{{$\in$}}1
             {→}{{$\rightarrow$}}1 {◁}{{$\triangleleft$}}1
             {≃}{{$\simeq$}}1
             {⊤}{{$\top$}}1 {⁅}{{$[\![$}}2 {⁆}{{$]\!]$}}2
             {⁻¹}{{$^{-1}$}}2}

\definecolor{promptbarcolor}{gray}{0.38}
\definecolor{promptframecolor}{gray}{0.38}
\newtcblisting{promptbox}[1]{%
    enhanced, breakable,
    colback=white,
    colframe=promptframecolor,
    coltitle=white,
    colbacktitle=promptbarcolor,
    fonttitle=\small\bfseries,
    title={#1},
    listing only,
    listing options={style=prompt},
    boxrule=1pt,
    arc=2pt, outer arc=2pt,
    left=0.5mm, right=0.5mm, top=1mm, bottom=1mm,
    toptitle=0.6mm, bottomtitle=0.6mm, lefttitle=0.5mm}

\definecolor{okgreen}{HTML}{18794E}
\definecolor{failred}{HTML}{B42318}

\newcommand{\casetag}[1]{%
  \par\smallskip\noindent
  {\bfseries\footnotesize #1}%
  \par\smallskip\noindent}

\newcommand{\statusok}[1]{\textcolor{okgreen}{\ding{51}}~\textbf{#1}}
\newcommand{\statusfail}[1]{\textcolor{failred}{\ding{55}}~\textbf{#1}}
\newcommand{\caseverdict}[2][]{\tcbsubtitle[colback=gray!12, coltitle=black, #1]{#2}}

\newtcolorbox{caseboxlong}[1]{%
    enhanced, breakable,
    colback=white,
    colframe=promptframecolor,
    coltitle=white,
    colbacktitle=promptbarcolor,
    fonttitle=\small\bfseries,
    fontupper=\small,
    title={#1},
    boxrule=1pt,
    arc=2pt, outer arc=2pt,
    left=2mm, right=2mm, top=1.5mm, bottom=1.5mm,
    toptitle=0.6mm, bottomtitle=0.6mm, lefttitle=1mm,
    subtitle style={colback=gray!20, coltitle=black, colupper=black,
                    fonttitle=\small\bfseries, boxrule=0pt, sharp corners,
                    left=1.5mm, top=0.9mm, bottom=0.9mm}}

\title{MathForm: Scaling Mathematical Autoformalization with Knowledge Retrieval and Verification-Guided Refinement}

\author{%
Lushi Pu\textsuperscript{\rm 1}, 
Weiming Zhang\textsuperscript{\rm 2},
Xinheng Xie\textsuperscript{\rm 1},
Zixuan Fu\textsuperscript{\rm 2},
Bingxiang He\textsuperscript{\rm 2},
Hengyu Zhao\textsuperscript{\rm 1},\\
\textbf{
Hongya Lyu\textsuperscript{\rm 1}, 
Xin Li\textsuperscript{\rm 1},
Jie Zhou\textsuperscript{\rm 1},
Yudong Wang\textsuperscript{\rm 2}$^{\dagger}$
} \\
\textsuperscript{\rm 1}ModelBest Inc. \quad
\textsuperscript{\rm 2}Tsinghua University ~~~~ \\
\texttt{pulushi@modelbest.cn} \quad
\texttt{yudongwang@tsinghua.edu.cn}
}

\newcommand\blfootnote[1]{%
\begingroup
\renewcommand\thefootnote{}\footnote{#1}%
\addtocounter{footnote}{-1}%
\endgroup
}

\begin{document}

\setlength{\headwidth}{\textwidth}

\maketitle
\blfootnote{$\dagger$ Corresponding authors.}

\thispagestyle{fancy} 

\vspace{-2em}

\begin{abstract}
Autoformalization is commonly framed as translating natural-language mathematical statements into machine-verifiable formal languages such as Lean~4.
However, faithful formalization requires more than translation. Models must map mathematical concepts to the complex hierarchy of types and definitions in formal libraries such as Mathlib, while ensuring that generated statements preserve the meaning of the source propositions. Existing approaches struggle because they rely heavily on the model's parametric memory for library-specific knowledge, while common data construction pipelines often resort to filtering single-pass outputs and lack mechanisms for feedback-driven revision.
To address these challenges, we introduce \textit{\textbf{MathForm}}, an autoformalization framework for constructing verified training data through Mathlib knowledge retrieval and verification-guided iterative refinement. Before generation, a retrieval planner gathers relevant definitions and existing formalizations from Mathlib to guide the formalization generator. Generated statements are then revised using compiler diagnostics and semantic-consistency feedback.
Using this framework, we construct \textit{\textbf{FormalVerse}}, a Lean~4 dataset containing approximately 367K verified examples across diverse mathematical domains and sources. We then train \textit{\textbf{MathForm}}-8B through supervised fine-tuning followed by reinforcement learning.
Across six benchmarks, MathForm-8B achieves average Pass@8 rates of 88.06\% under Syntax Check (SC) and 72.37\% under Consistency Check (CC), outperforming multiple specialized 32B autoformalizers. On the challenging FATE-H and FATE-X subsets, it attains CC pass rates of 63\% and 37\%, exceeding the strongest specialized baselines in both cases.
\end{abstract}

\section{Introduction}

Recent advances in large language models (LLMs) have substantially improved formal theorem proving, enabling systems such as AlphaProof~\citep{hubert2026olympiad}, DeepSeek-Prover-V2~\citep{ren2025deepseekproverv2advancingformalmathematical}, and Goedel-Prover-V2~\citep{lin2025goedelproverv2scalingformaltheorem} to generate sophisticated Lean~4~\citep{moura2021lean} proofs for formally stated problems.
However, scaling these systems further requires large and diverse corpora of machine-checkable statements and proofs, which remain scarce.
A central bottleneck is that much mathematical knowledge exists only in natural language, and encoding it manually in a formal language demands both precise mathematical interpretation and considerable expertise with the proof assistant itself.
Autoformalization addresses this bottleneck by translating natural-language mathematical statements into machine-checkable representations such as Lean~4 code, providing a scalable way to unlock these informal resources.

Despite notable progress in autoformalization, existing approaches face two methodological limitations.
First, prevailing methods~\citep{wang2024theoremllamatransforminggeneralpurposellms, gao2025heraldnaturallanguageannotated, xuejun2026mathesis} rely primarily on the mathematical and programming knowledge stored in parametric memory.
However, formalization in Lean requires not only an understanding of the underlying mathematical semantics but also familiarity with Mathlib's~\citep{The_mathlib_Community_2020} definitions, type system, notational conventions, existing structures, and common formalization patterns.
Such knowledge is highly specific and evolves as the library develops, which makes it difficult to internalize fully in model parameters. Models therefore tend to misuse existing definitions, invoke lemmas that do not exist, or produce expressions that are valid yet depart from library conventions.
Second, many existing pipelines for constructing autoformalization data~\citep{wang2025kiminaproverpreviewlargeformal, wu2025stepfunformalizerunlockingautoformalizationpotential, yu2025formalmathbenchmarkingformalmathematical} adopt a Best-of-$N$ (BoN) strategy, in which the model samples a large pool of candidates that a discriminator then filters post hoc.
Such pipelines can improve the quality of the resulting data, but they merely select from the model's existing output distribution, since every candidate comes from single-pass generation. Moreover, the discriminator only accepts or rejects a candidate as a whole, so it cannot indicate where a semantic deviation occurs or how to repair it.
The difficulty of the data is therefore capped at the model's current single-pass capability.
Together, these two limitations leave existing datasets and models concentrated on competition-style algebra and number theory~\citep{zheng2022miniff}, while statements with complex preconditions and domains that demand deeper library knowledge, such as abstract algebra~\citep{jiang2026fate}, remain underrepresented.

\begin{figure}[tb]
    \centering
    \includegraphics[width=0.85\textwidth]{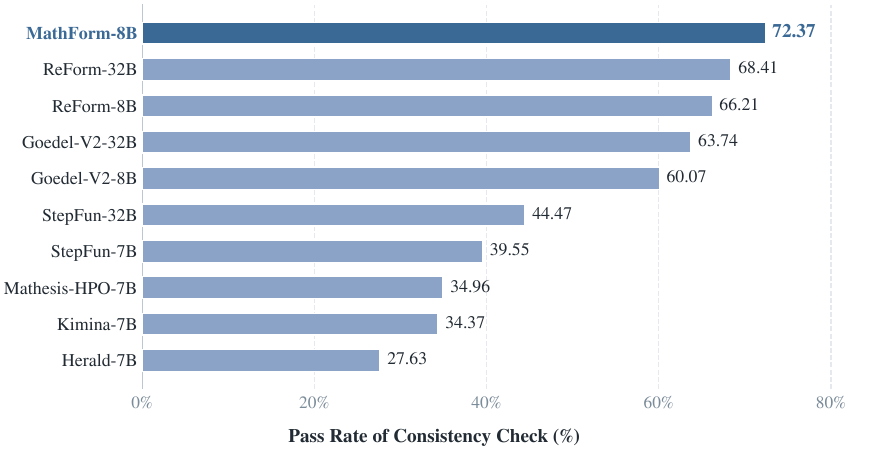}
    \caption{Macro-average Pass@8 (\%) across six benchmarks among specialized autoformalizers. \textsc{MathForm}-8B achieves the strongest overall performance within this category despite its smaller model size.}
    \label{fig:pass8-comparison}
\end{figure}

We argue that both limitations stem from treating autoformalization as a one-shot translation to be judged after the fact, whereas faithful formalization is better understood as a knowledge-grounded process that converges through repeated verification.
Because a single mathematical object may correspond to a complex hierarchy of types and definitions in Mathlib, a plausible-looking statement can compile and still strengthen a condition or drop a key assumption. Such a deviation renders the statement of little use downstream.

To this end, we propose \textsc{MathForm}, a framework for autoformalization data construction and model training that organizes library knowledge retrieval, automated verification, and iterative refinement into a closed loop.
Before generation, retrieval supplies definitions and existing formalizations from Mathlib, so the model need not recall library knowledge from parameters alone.
Compiler diagnostics and semantic-consistency judgments then locate what is wrong with a candidate rather than merely rejecting it, and iterative refinement turns that feedback into concrete revisions.
The ceiling on data difficulty is then set by the pipeline as a whole rather than by single-pass generation, so it can formalize statements that a single pass fails to handle, extending the training data to harder statements and more advanced mathematical domains.
Training on such data compresses the capability of the whole pipeline into the model's single-pass generation, realizing a form of data--model co-evolution~\citep{wang2026datasciencetechnologyagi}.

Using this framework, we construct \textsc{FormalVerse}, a Lean~4 dataset of approximately 367K verified examples spanning diverse mathematical domains and problem sources (Appendix Figure~\ref{fig:category-distribution}), and reconstruct a clean formalization trajectory for each example.
Building on these data, we train \textsc{MathForm}-8B through supervised fine-tuning (SFT) followed by reinforcement learning (RL).
On FormalMATH-Lite~\citep{yu2025formalmathbenchmarkingformalmathematical}, DeepSeek-ProverBench~\citep{ren2025deepseekproverv2advancingformalmathematical}, CombiBench~\citep{liu2026combibench}, and FATE~\citep{jiang2026fate}, it outperforms existing models of comparable size under Pass@8 and matches or exceeds substantially larger specialized autoformalizers on several challenging test sets (Figure~\ref{fig:pass8-comparison}).
Ablations further show that knowledge retrieval, automated verification, and reinforcement learning each contribute to this result.

In summary, our main contributions are as follows:
\begin{itemize}
    \item We introduce \textsc{MathForm}, a knowledge-augmented autoformalization framework that combines retrieval planning with compiler- and semantics-guided iterative refinement, enabling reliable natural-language-to-Lean data construction beyond reliance on parametric memory.
    \item We construct \textsc{FormalVerse}, a large-scale Lean~4 autoformalization dataset containing approximately 367K verified examples across diverse data sources, mathematical domains, and problem types.
    \item Using a straightforward training recipe consisting solely of SFT and RL, we train \textsc{MathForm}-8B, which achieves average pass rates of 88.06\% under Syntax Check and 72.37\% under Consistency Check across six benchmarks, outperforming multiple specialized 32B autoformalizers.
\end{itemize}

\begin{figure}[tb]
    \centering
    \includegraphics[width=0.95\textwidth]{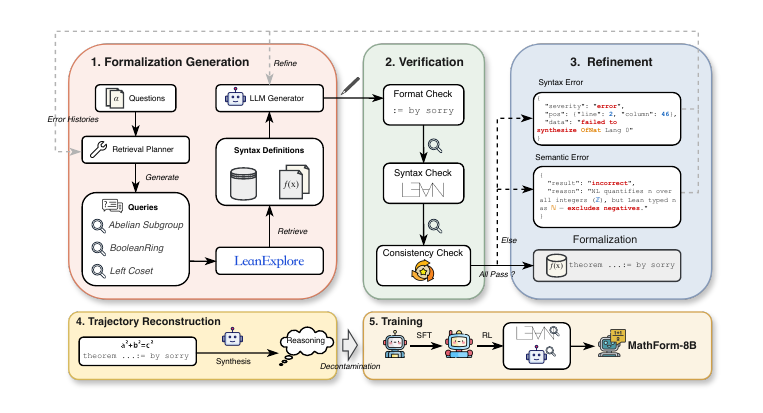}
    \caption{Overview of the \textsc{MathForm} data construction and training pipeline. The system combines Mathlib knowledge retrieval, compilation and semantic verification, and iterative refinement to generate reliable formal data, followed by trajectory reconstruction and training of \textsc{MathForm}-8B.}
    \label{fig:data-pipeline}
\end{figure}

\section{Related Work}

\subsection{Autoformalization}
Autoformalization aims to translate mathematical problems expressed in natural language into machine-verifiable formal code, and constitutes a key subtask in formal theorem proving.
Whereas early rule-based methods were limited in both accuracy and coverage, recent research has shifted toward LLM-based approaches to autoformalization.
Systems such as TheoremLlama~\citep{wang2024theoremllamatransforminggeneralpurposellms}, Herald~\citep{gao2025heraldnaturallanguageannotated}, Kimina-Autoformalizer~\citep{wang2025kiminaproverpreviewlargeformal}, and Mathesis~\citep{xuejun2026mathesis} train end-to-end natural-language-to-formal-language (NL-to-FL) autoformalization models.
Subsequent systems, including StepFun-Formalizer~\citep{wu2025stepfunformalizerunlockingautoformalizationpotential}, Goedel-Formalizer-V2~\citep{lin2025goedelproverv2scalingformaltheorem}, ATF~\citep{guo2025autoformalizertoolfeedback}, and ReForm~\citep{chen2026reformreflectiveautoformalizationprospective}, incorporate verifier feedback to train autoformalizers capable of explicit reasoning and refinement.
Meanwhile, RAutoformalizer~\citep{liu2025rethinking}, DRIFT~\citep{zhang2026driftdecomposeretrieveillustrate}, and Aria~\citep{wang2026ariaagentretrievaliterative} explore retrieval-augmented frameworks for autoformalization.
These retrieval-centric approaches, however, primarily target per-instance formalization at inference time and have not been designed for large-scale, end-to-end autoformalization data construction.

\subsection{Datasets for Formal Mathematical Reasoning}
High-quality formal data have long remained a scarce resource for research on formal mathematical reasoning.
MiniF2F~\citep{zheng2022miniff} and ProofNet~\citep{azerbayev2023proofnetautoformalizingformallyproving} formalize Olympiad-level problems and undergraduate mathematical theorems as Lean statements, providing important benchmarks for autoformalization and formal theorem proving.
More recently, CombiBench~\citep{liu2026combibench}, FATE~\citep{jiang2026fate}, and MA-ProofBench~\citep{pu2026maproofbenchtwotieredevaluationllms} extend formal evaluation to combinatorics, abstract algebra, and mathematical analysis, domains that earlier benchmarks cover only sparsely.
LeanDojo~\citep{yang2023leandojo} subsequently introduced an interactive Lean proving environment built upon Mathlib.
More recent efforts have constructed large-scale Lean~4 datasets for model training, including Lean Workbook~\citep{ying2024lean}, Lean-GitHub~\citep{wu2024leangithubcompilinggithublean}, NuminaMath-LEAN~\citep{wang2025kiminaproverpreviewlargeformal}, and FineLeanCorpus~\citep{peng-etal-2026-criticlean}.
However, their semantic quality and domain coverage remain limited.
Our approach instead combines efficient knowledge augmentation, automated verification, and iterative refinement to construct higher-quality formal data and thereby improve model autoformalization capabilities.

\section{Method}
\label{sec:method}

This section presents the data construction framework and model training procedure of \textsc{MathForm}. As illustrated in Figure~\ref{fig:data-pipeline}, Section~\ref{sec:autoformalization-framework} describes the data construction framework, covering problem collection and normalization, knowledge retrieval and formalization generation, verification-guided iterative refinement, and trajectory reconstruction followed by data decontamination. Section~\ref{sec:training-mathform} then presents the training of \textsc{MathForm}-8B.

\begin{figure}[t]
    \centering
    \includegraphics[width=0.55\textwidth]{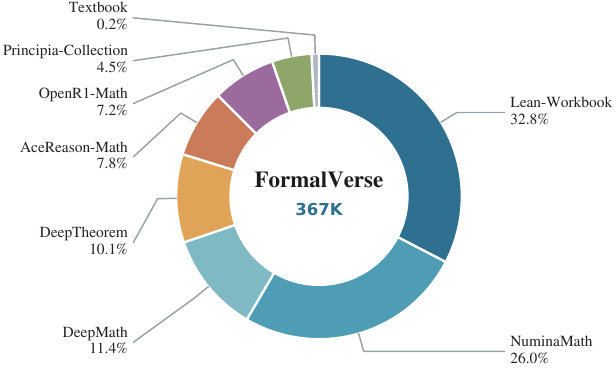}
    \caption{Distribution of natural-language problem sources in \textsc{FormalVerse}.}
    \label{fig:source-distribution}
\end{figure}

\subsection{Autoformalization Framework}
\label{sec:autoformalization-framework}

\subsubsection{Problem Collection and Normalization}
We first collect candidate problems from a diverse set of natural-language mathematics datasets, including DeepTheorem~\citep{zhang2025deeptheoremadvancingllmreasoning}, NuminaMath~\citep{li2024numinamath}, AceReason-Math~\citep{chen2025acereasonnemotron}, Lean Workbook~\citep{ying2024lean}, Principia-Collection~\citep{aggarwal2026reasoningmathematicalobjectsonpolicy}, DeepMath~\citep{he2026deepmathk}, and OpenR1-Math~\citep{openr1}, and further supplement them with theorems and exercises drawn from classical mathematics textbooks.
We then filter out non-mathematical content, purely numerical computation exercises, and problems that cannot be naturally expressed as theorem statements.
Problems containing redundant answer-format instructions or extraneous context are rewritten, giving a normalized pool of natural-language problems. Figure~\ref{fig:source-distribution} shows the source composition of the final \textsc{FormalVerse} dataset.

\subsubsection{Knowledge Retrieval and Formalization Generation}
We build a knowledge-augmented autoformalization pipeline composed of a \textit{\textbf{Retrieval Planner}} and a \textit{\textbf{Formalization Generator}}. To keep inference fast and inexpensive, we use gpt-oss-120b~\citep{openai2025gptoss120bgptoss20bmodel} to drive both modules.
Given a natural-language mathematical statement, the retrieval planner analyzes the mathematical objects, relations, and type constraints it involves, and judges whether additional Mathlib knowledge is needed.
When this is the case, the planner issues a few targeted queries for the key concepts and collects relevant definitions, theorems, notations, and existing formalizations from Mathlib through LeanExplore~\citep{asher2025leanexploresearchenginelean}, which is set to return the top-$2$ results per query.
The formalization generator then conditions on both the original statement and these retrieved results, identifies the corresponding Mathlib types and definitions, and produces the Lean~4 formal statement.
Separating retrieval planning from code generation allows the pipeline to draw on library knowledge only where it is needed, which reduces its reliance on parametric memory and improves agreement with canonical Mathlib representations.

\subsubsection{Verification-Guided Iterative Refinement}
\begin{figure}[tb]
    \centering
    \includegraphics[width=0.65\textwidth]{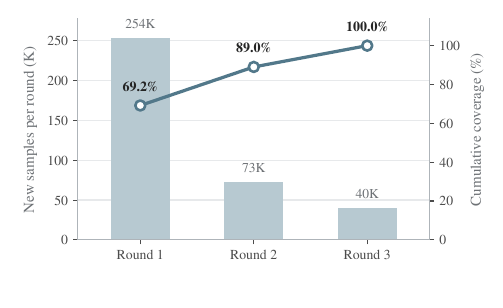}
    \caption{Numbers of natural-language-to-formal-language pairs accepted in successive refinement rounds. Later rounds contribute an additional 31.0\% of all retained pairs.}
    \label{fig:refinement-round-curve}
\end{figure}
Given a natural-language statement, the retrieved context, and feedback from any previous failed attempt, the formalization generator produces a candidate Lean~4 formalization.
A Format Check first discards outputs containing proof steps, tactics, solution procedures, or other content beyond the formal statement itself.
The remaining candidates are then compiled with Lean~4.
Compilation failures are recorded with the corresponding compiler diagnostics, which identify issues such as syntax errors, undeclared identifiers, missing dependencies, or type mismatches.

Successfully compiled candidates then undergo a semantic consistency check that assesses whether the generated Lean~4 statement faithfully captures the semantics of the original natural-language statement, with QwQ-32B~\citep{qwq32b} serving as the judge.
This check flags errors such as omitted assumptions, strengthened or weakened conditions, incorrect quantifier order, extraneous constraints, inappropriate mathematical objects, and mismatched conclusions.
A sample that fails either check proceeds to the next round together with the corresponding feedback, and Mathlib retrieval is re-triggered when additional context is needed.
Each sample undergoes at most three rounds, and generation stops as soon as a candidate passes both checks; samples that never pass are discarded.
This adaptive, failure-driven schedule concentrates additional computation on unresolved cases. Whereas Best-of-$N$ spends a fixed budget on independent samples, \textsc{MathForm} turns verification signals into corrective guidance rather than using them only for post-hoc selection.
As shown in Figure~\ref{fig:refinement-round-curve}, the first round yields about 69\% of all retained pairs, and the second and third rounds add roughly 20\% and 11\%, respectively. Later rounds therefore recover 31\% of the data that single-pass generation alone would not have produced.
A worked example is provided in Appendix~\ref{sec:case-pipeline}.

\subsubsection{Trajectory Reconstruction}

The preceding pipeline produces a large collection of verified natural-language-to-formal-language (NL-FL) pairs.
However, the trajectories that produced these pairs span multiple rounds and interleave retrieved context, compiler errors, semantic feedback, and failed attempts, which makes them unsuitable as training targets.
The native reasoning traces of the model are also unnecessarily verbose: even when the prompt asks only for a formal statement, it sometimes deliberates repeatedly over whether to produce a proof or attempts to solve the problem outright. This behavior likely reflects the fact that reasoning models are trained predominantly to prove or solve mathematical problems rather than to formalize them.
We therefore retrospectively synthesize a clean, structured formalization trajectory for each verified pair.
Specifically, given a natural-language mathematical statement and its Lean~4 formalization, we ask the model to reconstruct the intermediate analysis that maps one to the other, while explicitly excluding proof strategies, tactic selection, and problem-solving procedures.
Each resulting training example thus consists of a natural-language statement, a formalization trajectory, and verified Lean~4 code.
The trajectory reconstruction prompt is provided in Appendix~\ref{sec:prompts}.

\subsubsection{Data Decontamination}

Finally, we decontaminate the training data against all evaluation benchmarks used in Section~\ref{sec:experiments}, removing any training example that shares at least one 13-gram with an evaluation example.
The resulting dataset, \textsc{FormalVerse}, contains approximately 367K verified NL-FL pairs.

\subsection{Training \textsc{MathForm}-8B}
\label{sec:training-mathform}

\subsubsection{Supervised Fine-Tuning}

We perform supervised fine-tuning on Qwen3-8B~\citep{yang2025qwen3technicalreport} with \textsc{FormalVerse} using the LLaMA-Factory framework~\citep{zheng2024llamafactoryunifiedefficientfinetuning}, training the model to identify mathematical objects, logical structures, variable dependencies, and implicit type constraints in natural-language statements and to generate the corresponding Lean~4 formalizations.
This stage yields \textsc{MathForm}-8B-SFT, which already exhibits strong formalization ability and serves as the initialization for subsequent reinforcement learning.

\subsubsection{RL Data Selection}
The RL data are drawn from statements that never passed validation during iterative refinement, that is, the problems left unsolved by the construction pipeline. We sample approximately 20,000 such candidates.
We then apply offline difficulty filtering, keeping statements that are hard enough to provide informative optimization signals yet remain within reach of \textsc{MathForm}-8B-SFT.
Statements whose phrasing admits multiple reasonable formalizations are also removed so that the binary reward remains well defined.
This procedure yields an RL dataset of 3,000 examples.

\subsubsection{Reward Function}
After supervised fine-tuning, we further optimize \textsc{MathForm}-8B-SFT with Decoupled Clip and Dynamic sAmpling Policy Optimization (DAPO)~\citep{yu2025dapoopensourcellmreinforcement} using the verl framework~\citep{Sheng_2025}.
For each natural-language statement $x$, we sample $G$ candidate formalizations $\{y_i\}_{i=1}^{G}$ from the previous policy $\pi_{\theta_{\mathrm{old}}}$ and optimize the following token-level objective:
\begin{equation}
J_{\mathrm{DAPO}}(\theta)
= \mathbb{E}\Bigg[
\frac{1}{\sum_{i=1}^{G}|y_i|}
\sum_{i=1}^{G}\sum_{t=1}^{|y_i|}
\min\Big(
\rho_{i,t}A_i,\;
\operatorname{clip}(\rho_{i,t},
1-\epsilon_{\mathrm{low}},
1+\epsilon_{\mathrm{high}})A_i
\Big)\Bigg].
\label{eq:dapo-objective}
\end{equation}
Here, $\rho_{i,t}$ denotes the token-level policy ratio and $A_i$ the group-normalized advantage. We adopt Clip-Higher and dynamic sampling, retaining only groups that contain both successful and unsuccessful candidates.

We use a binary reward that jointly considers compilation success and semantic consistency.
Let $C(y)$ denote Lean~4 compilation success and $S(x,y)$ denote the semantic-consistency judgment produced by gpt-oss-20b for a compilable formalization.
A candidate receives a positive reward only when both checks succeed:
\begin{equation}
r(x,y)=
\begin{cases}
1, & C(y)=1 \ \text{and}\ S(x,y)=1,\\
0, & \text{otherwise}.
\end{cases}
\label{eq:verification-reward}
\end{equation}
This reward directly aligns optimization with the two core objectives of autoformalization, namely compilability and semantic fidelity. Full training details are provided in Appendix~\ref{sec:appendix}.

\section{Experiments}
\label{sec:experiments}

\begin{table}[tb]
\centering
\small
\setlength{\tabcolsep}{4pt}
\resizebox{\textwidth}{!}{%
\begin{tabular}{@{}l*{14}{c}@{}}
\toprule
& \multicolumn{2}{c}{\textbf{AVG}}
& \multicolumn{2}{c}{\textbf{FormalMATH}}
& \multicolumn{2}{c}{\textbf{ProverBench}}
& \multicolumn{2}{c}{\textbf{CombiBench}}
& \multicolumn{2}{c}{\textbf{FATE-M}}
& \multicolumn{2}{c}{\textbf{FATE-H}}
& \multicolumn{2}{c}{\textbf{FATE-X}} \\
\cmidrule(lr){2-3}\cmidrule(lr){4-5}\cmidrule(lr){6-7}
\cmidrule(lr){8-9}\cmidrule(lr){10-11}\cmidrule(lr){12-13}
\cmidrule(lr){14-15}
\textbf{Model}
& \textbf{SC} & \textbf{CC}
& \textbf{SC} & \textbf{CC}
& \textbf{SC} & \textbf{CC}
& \textbf{SC} & \textbf{CC}
& \textbf{SC} & \textbf{CC}
& \textbf{SC} & \textbf{CC}
& \textbf{SC} & \textbf{CC} \\
\midrule
\multicolumn{15}{@{}l}{\textit{Specialized Autoformalizers}} \\
\addlinespace[1pt]
Herald Translator-7B
& 64.12 & 27.63 & 95.29 & 47.76 & 78.74 & 37.36
& 77.00 & 5.00 & 70.67 & 54.67 & 42.00 & 15.00 & 21.00 & 6.00 \\
Kimina-Autoformalizer-7B
& 73.20 & 34.37 & \underline{99.29} & 76.24 & 96.55 & 56.32
& \underline{95.00} & 16.00 & 77.33 & 44.67 & 43.00 & 8.00 & 28.00 & 5.00 \\
Mathesis-HPO-7B
& 76.20 & 34.96 & 99.06 & 79.29 & 97.13 & 59.77
& \textbf{96.00} & 15.00 & 84.00 & 48.67 & 50.00 & 4.00 & 31.00 & 3.00 \\
StepFun-Formalizer-7B
& 58.12 & 39.55 & 97.41 & 81.41 & 89.66 & 59.20
& 79.00 & 28.00 & 60.67 & 52.67 & 17.00 & 12.00 & 5.00 & 4.00 \\
StepFun-Formalizer-32B
& 63.65 & 44.47 & 99.06 & 85.88 & 92.53 & 64.94
& 86.00 & 32.00 & 71.33 & 60.00 & 23.00 & 17.00 & 10.00 & 7.00 \\
Goedel-Formalizer-V2-8B
& 78.24 & 60.08 & 98.82 & 94.12 & 98.28 & 89.66
& 89.00 & 42.00 & 87.33 & 82.67 & 62.00 & 44.00 & 34.00 & 8.00 \\
Goedel-Formalizer-V2-32B
& 78.28 & 63.74 & 99.06 & 94.59 & 98.28 & 92.53
& 91.00 & \underline{49.00} & 89.33 & 85.33 & 63.00 & 48.00 & 29.00 & 13.00 \\
ReForm-8B
& 81.76 & 66.21 & 99.06 & 94.12 & \underline{98.85} & 90.80
& 86.00 & 47.00 & 94.67 & \underline{91.33} & 67.00 & 53.00 & 45.00 & 21.00 \\
ReForm-32B
& 81.61 & \underline{68.41} & 99.06 & \textbf{95.53} & 98.28 & \underline{94.25}
& 93.00 & \textbf{55.00} & 91.33 & 88.67 & 69.00 & 52.00 & 39.00 & \underline{25.00} \\
\midrule
\multicolumn{15}{@{}l}{\textit{Ours}} \\
\addlinespace[1pt]
\rowcolor{oursrow}
\textsc{MathForm}-8B-SFT
& \underline{84.38} & 66.53 & \underline{99.29} & 91.06 & \textbf{100.00} & 90.80
& 83.00 & 43.00 & \underline{98.00} & \underline{91.33} & \underline{80.00} & \underline{58.00} & \underline{46.00} & \underline{25.00} \\
\rowcolor{oursrow}
\textbf{\textsc{MathForm}-8B}
& \textbf{88.06} & \textbf{72.37} & \textbf{100.00} & \underline{95.06}
& \textbf{100.00} & \textbf{94.83} & 93.00 & 47.00
& \textbf{99.33} & \textbf{97.33} & \textbf{82.00} & \textbf{63.00}
& \textbf{54.00} & \textbf{37.00} \\
\bottomrule
\end{tabular}%
}
\caption{Pass@8 pass rates (\%) under Syntax Check (SC) and Consistency Check (CC) for specialized autoformalizers on six benchmarks. AVG is the equally weighted macro-average across all six benchmarks. For each column, the best result is shown in \textbf{bold} and the second best is \underline{underlined}.}
\label{tab:main-results}
\end{table}

\subsection{Experimental Settings}

\subsubsection{Benchmarks}
We evaluate on six benchmarks: FormalMATH-Lite~\citep{yu2025formalmathbenchmarkingformalmathematical}, DeepSeek-ProverBench (ProverBench)~\citep{ren2025deepseekproverv2advancingformalmathematical}, CombiBench~\citep{liu2026combibench}, and the recently introduced and highly challenging FATE-M, FATE-H, and FATE-X~\citep{jiang2026fate}.
Together, these benchmarks cover competition mathematics, combinatorics, and algebraic reasoning ranging from elementary abstract algebra to advanced commutative algebra, homological algebra, and foundations of algebraic geometry.

\subsubsection{Baselines}
We compare \textsc{MathForm}-8B with a range of models, including specialized autoformalizers and general-purpose LLMs.
The specialized autoformalizers include Herald Translator-7B~\citep{gao2025heraldnaturallanguageannotated}, Kimina-Autoformalizer-7B~\citep{wang2025kiminaproverpreviewlargeformal}, Mathesis-HPO-7B~\citep{xuejun2026mathesis}, StepFun-Formalizer-7B/32B~\citep{wu2025stepfunformalizerunlockingautoformalizationpotential}, Goedel-Formalizer-V2-8B/32B~\citep{lin2025goedelproverv2scalingformaltheorem}, and ReForm-8B/32B~\citep{chen2026reformreflectiveautoformalizationprospective}.
The general-purpose LLMs include DeepSeek-V4-Pro~\citep{deepseekai2026deepseekv4highlyefficientmilliontoken}, Qwen3.7-Plus~\citep{qwen37plus}, Qwen3-8B/32B/235B-A22B-Thinking-2507~\citep{yang2025qwen3technicalreport}, and DeepSeek-R1-0528-Qwen3-8B~\citep{Guo_2025}; their complete results are reported in Appendix Table~\ref{tab:general-model-results}.
Table~\ref{tab:main-results} focuses on specialized autoformalizers to directly compare task-specific capabilities across model scales and training strategies.

\subsubsection{Evaluation}
We report Pass@$k$ rates~\citep{chen2021evaluatinglargelanguagemodels} under two criteria: Syntax Check (SC) and Consistency Check (CC).
For a source statement $x$ and its $k$ candidate formalizations $\{y_i\}_{i=1}^{k}$, let $C(y_i)$ be the binary indicator of compilation success and $S(x,y_i)$ the binary indicator of semantic consistency. The per-instance pass indicators under the two criteria are defined as
\begin{equation}
\begin{aligned}
\operatorname{SC@k}(x)
&=\max_{1\leq i\leq k} C(y_i),\\
\operatorname{CC@k}(x)
&=\max_{1\leq i\leq k} C(y_i)S(x,y_i).
\end{aligned}
\label{eq:pass-k-metrics}
\end{equation}
Here, $S$ is evaluated only for candidates that compile successfully; the final SC and CC pass rates are obtained by averaging the corresponding indicators over all test instances.
At inference time, we set $k=8$ and sample candidate formalizations for each test statement with a temperature of 0.6.
For compilation validation, we use Kimina Lean Server~\citep{santos2025kiminaleanserverhighperformance} as the Lean~4 backend to support efficient, large-scale checking of formal code.
For semantic-consistency evaluation, we adopt an LLM-as-a-Judge protocol, with gpt-oss-120b serving as the judge under the \texttt{high} reasoning-effort setting.
The complete semantic consistency check prompt and the per-model inference prompts are provided in Appendix~\ref{sec:prompts}.
To assess the robustness of semantic-consistency evaluation, we further compare multiple judge models in our subsequent analysis (Table~\ref{tab:judge-reliability}).

\begin{figure}[tb]
    \centering
    \includegraphics[width=0.65\textwidth]{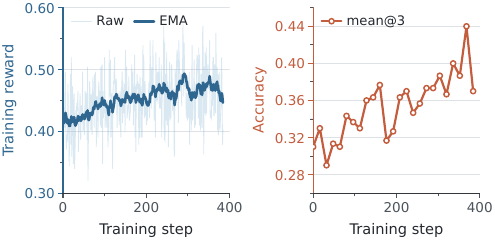}
    \caption{Training dynamics during reinforcement learning. The left panel shows the training reward and its exponential moving average (EMA), while the right panel reports the Mean@3 pass rate on FATE-H over the course of training.}
    \label{fig:training-dynamics}
\end{figure}

\subsection{Main Results}

\subsubsection{An 8B Model Surpasses 32B Specialized Autoformalizers}
As shown in Table~\ref{tab:main-results}, \textsc{MathForm}-8B achieves the best average SC and CC pass rates among specialized autoformalizers, reaching 88.06\% and 72.37\%, respectively.
Compared with the strongest specialized baseline, ReForm-32B (81.61/68.41), these results represent absolute gains of 6.45 and 3.96 percentage points.
Part of this margin comes from the RL stage: relative to \textsc{MathForm}-8B-SFT, reinforcement learning raises the average SC pass rate from 84.38\% to 88.06\% and the average CC pass rate from 66.53\% to 72.37\%. The larger CC gain indicates that verification-driven reinforcement learning improves semantic alignment beyond compilability.
Human evaluation on FATE-M and FATE-H confirms this ranking, where \textsc{MathForm}-8B achieves the highest SC and human-assessed CC pass rates among all compared models (Appendix Table~\ref{tab:human-evaluation}).

\subsubsection{Gains Concentrate on High-Abstraction Statements}
On the relatively mature FormalMATH-Lite and ProverBench benchmarks, specialized models perform comparably, and \textsc{MathForm}-8B is on par with or slightly better than the strongest baseline.
On CombiBench, \textsc{MathForm}-8B matches ReForm-8B in CC.
The separation emerges on the FATE series: \textsc{MathForm}-8B attains CC pass rates of 97.33\%, 63.00\%, and 37.00\% on FATE-M, FATE-H, and FATE-X, exceeding the strongest specialized baseline on each subset by 6, 10, and 12 percentage points, with the advantage widening as the abstraction level increases.
Advanced algebra is precisely where formalization depends most on Mathlib's type hierarchy and existing formalizations, and this distribution of gains is consistent with the knowledge retrieval and verification-guided iterative refinement that underlie the construction of \textsc{FormalVerse}.
A representative case is provided in Appendix~\ref{sec:case-mathform}.

\subsection{Further Analysis}

\begin{table}[tb]
\centering
\small
\setlength{\tabcolsep}{8pt}
\begin{tabular}{@{}l*{8}{c}@{}}
\toprule
& \multicolumn{2}{c}{\textbf{AVG}}
& \multicolumn{2}{c}{\textbf{FATE-M}}
& \multicolumn{2}{c}{\textbf{FATE-H}}
& \multicolumn{2}{c}{\textbf{FATE-X}} \\
\cmidrule(lr){2-3}\cmidrule(lr){4-5}\cmidrule(lr){6-7}\cmidrule(lr){8-9}
\textbf{Method}
& \textbf{SC} & \textbf{CC}
& \textbf{SC} & \textbf{CC}
& \textbf{SC} & \textbf{CC}
& \textbf{SC} & \textbf{CC} \\
\midrule
\multicolumn{9}{@{}l}{\textit{gpt-oss-120b}} \\
\addlinespace[1pt]
Single
& 27.33 & 26.43 & 52.00 & 51.30 & 21.00 & 19.00 & 9.00 & 9.00 \\
BoN
& 42.67 & 41.10 & 70.00 & 69.30 & 40.00 & 37.00 & 18.00 & 17.00 \\
Feedback
& 42.57 & 40.77 & 68.70 & 67.30 & 41.00 & 38.00 & 18.00 & 17.00 \\
Retrieval
& 32.23 & 29.00 & 56.70 & 52.00 & 27.00 & 26.00 & 13.00 & 9.00 \\
\rowcolor{oursrow}
\textbf{\textsc{MathForm}}
& \textbf{49.67} & \textbf{48.00} & \textbf{76.00} & \textbf{74.00}
& \textbf{46.00} & \textbf{44.00} & \textbf{27.00} & \textbf{26.00} \\
\midrule
\multicolumn{9}{@{}l}{\textit{Qwen3-235B-A22B-Thinking-2507}} \\
\addlinespace[1pt]
Single
& 7.43 & 7.43 & 13.30 & 13.30 & 5.00 & 5.00 & 4.00 & 4.00 \\
BoN
& 18.77 & 18.77 & 33.30 & 33.30 & 14.00 & 14.00 & 9.00 & 9.00 \\
Feedback
& 28.77 & 28.77 & 51.30 & 51.30 & 22.00 & 22.00 & 13.00 & 13.00 \\
Retrieval
& 9.90 & 8.57 & 16.70 & 16.70 & 8.00 & 6.00 & 5.00 & 3.00 \\
\rowcolor{oursrow}
\textbf{\textsc{MathForm}}
& \textbf{37.57} & \textbf{36.23} & \textbf{60.70} & \textbf{58.70}
& \textbf{32.00} & \textbf{31.00} & \textbf{20.00} & \textbf{19.00} \\
\bottomrule
\end{tabular}
\caption{Ablation results for SC and CC pass rates (\%) of the refinement pipeline with two generators on the FATE series. Single denotes single-pass generation; BoN denotes Best-of-$N$; Feedback denotes feedback-only iteration; and Retrieval denotes retrieval-only single-pass generation. AVG is the equally weighted average over FATE-M, FATE-H, and FATE-X.}
\label{tab:refinement-ablation}
\end{table}

\begin{table}[tb]
\centering
\small
\setlength{\tabcolsep}{2.2pt}
\resizebox{\textwidth}{!}{%
\begin{tabular}{@{}l*{14}{c}@{}}
\toprule
& \multicolumn{2}{c}{\textbf{AVG}}
& \multicolumn{2}{c}{\textbf{FormalMATH}}
& \multicolumn{2}{c}{\textbf{ProverBench}}
& \multicolumn{2}{c}{\textbf{CombiBench}}
& \multicolumn{2}{c}{\textbf{FATE-M}}
& \multicolumn{2}{c}{\textbf{FATE-H}}
& \multicolumn{2}{c}{\textbf{FATE-X}} \\
\cmidrule(lr){2-3}\cmidrule(lr){4-5}\cmidrule(lr){6-7}
\cmidrule(lr){8-9}\cmidrule(lr){10-11}\cmidrule(lr){12-13}
\cmidrule(lr){14-15}
\textbf{Training Dataset}
& \textbf{SC} & \textbf{CC}
& \textbf{SC} & \textbf{CC}
& \textbf{SC} & \textbf{CC}
& \textbf{SC} & \textbf{CC}
& \textbf{SC} & \textbf{CC}
& \textbf{SC} & \textbf{CC}
& \textbf{SC} & \textbf{CC} \\
\midrule
NuminaMath-LEAN
& 66.24 & 41.49
& \underline{99.53} & \underline{85.18}
& \underline{96.55} & 72.41
& \underline{87.00} & 25.00
& 69.33 & 49.33
& 35.00 & 16.00
& 10.00 & 1.00 \\
FineLeanCorpus
& \textbf{78.25} & \underline{46.53}
& \textbf{100.00} & 84.47
& \textbf{98.85} & \underline{74.71}
& \textbf{96.00} & \underline{29.00}
& \underline{90.67} & \underline{68.00}
& \underline{53.00} & \underline{17.00}
& \textbf{31.00} & \underline{6.00} \\
\rowcolor{oursrow}
\textbf{\textsc{FormalVerse}}
& \underline{77.17} & \textbf{60.32}
& 98.82 & \textbf{90.59}
& \textbf{98.85} & \textbf{89.66}
& 78.00 & \textbf{36.00}
& \textbf{95.33} & \textbf{84.67}
& \textbf{62.00} & \textbf{46.00}
& \underline{30.00} & \textbf{15.00} \\
\bottomrule
\end{tabular}%
}
\caption{Pass@8 SC and CC pass rates (\%) of models trained on different datasets across six benchmarks. All models are initialized from Qwen3-8B and trained on 100K examples under the same trajectory reconstruction and training configuration. AVG is the equally weighted macro-average across all six benchmarks. The best result in each column is shown in \textbf{bold} and the second best is \underline{underlined}.}
\label{tab:training-corpus-comparison}
\end{table}

\subsubsection{Training Dynamics}
During training, we track the pass rate on the challenging FATE-H benchmark using Mean@3, the fraction of candidates passing the Consistency Check among three samples per statement, averaged over the benchmark.
As shown in Figure~\ref{fig:training-dynamics}, the average reward increases steadily throughout training, indicating that the model progressively learns to generate formalizations satisfying both compilation success and semantic consistency.
Meanwhile, Mean@3 on FATE-H improves from 0.30 to about 0.40, corresponding to an approximately 33\% relative gain.
The close agreement between the reward curve and the FATE-H pass rate suggests that the proposed verification-driven reward reliably reflects actual autoformalization performance and provides an effective learning signal.

\subsubsection{Ablation Study of the Refinement Pipeline}
To quantify the contributions of knowledge retrieval and verification-guided iterative refinement, we compare five generation configurations on FATE-M, FATE-H, and FATE-X: single-pass generation, budget-matched Best-of-$N$ independent sampling with $N=3$, feedback-only iteration without retrieval, retrieval-only single-pass generation, and the complete \textsc{MathForm} pipeline.
We conduct the experiments with both gpt-oss-120b and Qwen3-235B-A22B-Thinking-2507, the latter belonging to a different model family.
The ablation results for both generators are reported in Table~\ref{tab:refinement-ablation}.
For both generators, the complete \textsc{MathForm} pipeline achieves the best SC and CC pass rates across all FATE difficulty levels, indicating that its effectiveness is not tied to a particular generator.
Relative to single-pass generation, \textsc{MathForm} improves the average SC pass rate from 27.33\% to 49.67\% for gpt-oss-120b (+22.34 percentage points) and from 7.43\% to 37.57\% for the Qwen3-235B generator (+30.14 percentage points).
More importantly, compared with the strongest single-component configuration among retrieval, feedback iteration, and budget-matched sampling, the full pipeline yields additional gains of 7.00/6.90 and 8.80/7.46 percentage points in SC/CC pass rates for the two generators, respectively.
These consistent cross-model gains demonstrate that knowledge retrieval and verification-guided refinement are complementary and jointly improve the autoformalization quality of the data construction pipeline.

\subsubsection{Data Quality of FormalVerse}
To isolate the effect of training-data quality, we compare \textsc{FormalVerse}
with two recently released large-scale Lean~4 datasets,
NuminaMath-LEAN~\citep{wang2025kiminaproverpreviewlargeformal} and
FineLeanCorpus~\citep{peng-etal-2026-criticlean}. We randomly sample 100K
examples from each, apply the same trajectory-reconstruction procedure to all
three subsets, and use each resulting dataset to fine-tune Qwen3-8B under an
identical training configuration. The resulting models are evaluated on all
six benchmarks using the same Pass@8 protocol.
Under this controlled setting (Table~\ref{tab:training-corpus-comparison}), \textsc{FormalVerse} attains the highest average
CC pass rate of 60.32\%, exceeding FineLeanCorpus by 13.79 and
NuminaMath-LEAN by 18.83 percentage points, and ranks first in CC on every
benchmark. Its SC pass rate (77.17\%), by contrast, is comparable to that of
FineLeanCorpus (78.25\%), so the difference between the two corpora lies almost
entirely in semantic fidelity rather than compilability.
This is the dimension that matters for autoformalization, since a statement
that compiles but misstates the source proposition is of little use
downstream, and it is precisely what the verification-guided construction of
\textsc{FormalVerse} targets.

\begin{table}[tb]
\centering
\small
\setlength{\tabcolsep}{10pt}
\begin{tabular}{lrrrr}
\toprule
\textbf{Judge Model} & \textbf{Accuracy} & \textbf{Precision} & \textbf{Recall} & \textbf{F1} \\
\midrule
gpt-oss-120b
& 0.8917 & 0.8755 & 0.9133 & 0.8940 \\
QwQ-32B
& 0.8567 & 0.8367 & 0.8867 & 0.8609 \\
gpt-oss-20b
& 0.8500 & 0.8142 & 0.9067 & 0.8579 \\
\bottomrule
\end{tabular}
\caption{Judge-model reliability on the human-annotated semantic-consistency test set (Mean@3).}
\label{tab:judge-reliability}
\end{table}

\subsubsection{Reliability of Semantic Consistency Judgment}
To evaluate the reliability of judge models for semantic-consistency assessment, we construct an independent, human-annotated test set.
Specifically, we collect both accepted and rejected NL-FL pairs from the data construction stage and have human experts independently relabel them, resulting in a test set of 200 examples.
We evaluate the gpt-oss-120b judge used in our main evaluation against QwQ-32B and gpt-oss-20b on this test set. Accuracy, precision, recall, and F1, each averaged over three independent judging runs (Mean@3), are reported in Table~\ref{tab:judge-reliability}.
The gpt-oss-120b judge achieves the best results across all four metrics.
QwQ-32B attains an F1 score of 0.8609 and exhibits strong discriminative ability; because it belongs to a different model family from the gpt-oss models used for generation and reward evaluation, we use it during data construction to reduce the self-preference bias that can arise when the generator and the judge come from the same model family.
Although gpt-oss-20b is smaller, its precision approaches that of QwQ-32B, its recall is higher, and its inference is faster, making it better suited for reward computation during reinforcement learning.

\section{Conclusion}

We introduced \textsc{MathForm}, an autoformalization framework that combines knowledge retrieval with iterative refinement guided by compilation and semantic signals.
Using this framework, we constructed \textsc{FormalVerse}, a dataset of approximately 367K verified Lean~4 examples, and trained \textsc{MathForm}-8B through SFT and RL.
Across six benchmarks, \textsc{MathForm}-8B achieves average SC and CC pass rates of 88.06\% and 72.37\%, respectively, outperforming multiple specialized 32B autoformalizers and obtaining strong results on challenging FATE subsets.
Overall, the results validate the effectiveness of knowledge retrieval and verification-guided iterative refinement for autoformalization data construction, and show that high-quality training data can support competitive autoformalization performance in a compact model.
In future work, we plan to explore larger-scale test-time scaling methods to further strengthen formalization on complex problems.


\bibliographystyle{citation}
\bibliography{citation}

\clearpage
\appendix
\section{Implementation Details}
\label{sec:appendix}

\subsection{Category Distribution of \textsc{FormalVerse}}
\begin{figure}[htbp]
    \centering
    \includegraphics[width=0.65\textwidth]{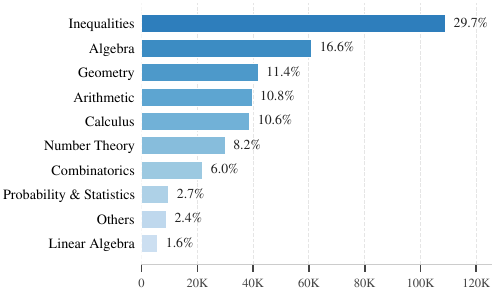}
    \caption{Category distribution of the natural-language mathematical problems in \textsc{FormalVerse}.}
    \label{fig:category-distribution}
\end{figure}

Figure~\ref{fig:category-distribution} presents the category distribution of the natural-language mathematical problems in \textsc{FormalVerse}.
The dataset covers ten mathematical categories, including inequalities, algebra, geometry, arithmetic, calculus, number theory, combinatorics, probability and statistics, and linear algebra.
This composition covers a wide variety of problem types, from competition mathematics to advanced topics, providing broad domain coverage for training autoformalizers.
\FloatBarrier
\FloatBarrier

\subsection{Training Hyperparameters}
The hyperparameters used for supervised fine-tuning and RL are summarized in Tables~\ref{tab:sft-hyperparameters} and~\ref{tab:rlvr-hyperparameters}, respectively.
Both training stages are conducted on 16 NVIDIA H100 80GB GPUs. All Lean compilation checks throughout this work use Lean 4.21.0.
\FloatBarrier
\begin{table}[htbp]
\centering
\small
\setlength{\tabcolsep}{6pt}
\begin{minipage}[t]{0.48\textwidth}
\centering
\begin{tabular}{@{}lr@{}}
\toprule
\textbf{Hyperparameter} & \textbf{Value} \\
\midrule
Maximum sequence length & 16,384 \\
Global batch size & 128 \\
Learning rate & $2.0\times10^{-5}$ \\
Epochs & 3 \\
LR scheduler & Cosine \\
Warmup ratio & 0.1 \\
Precision & \texttt{bf16} \\
\bottomrule
\end{tabular}
\caption{Supervised fine-tuning hyperparameters.}
\label{tab:sft-hyperparameters}
\end{minipage}
\hfill
\begin{minipage}[t]{0.48\textwidth}
\centering
\begin{tabular}{@{}lr@{}}
\toprule
\textbf{Hyperparameter} & \textbf{Value} \\
\midrule
RL algorithm & DAPO \\
Learning rate & $1.0\times10^{-6}$ \\
Clipping bounds (lower/upper) & 0.8 / 1.28 \\
Training batch size & 32 \\
PPO mini-batch size & 32 \\
PPO micro-batch size per GPU & 2 \\
Rollouts per prompt & 8 \\
Rollout temperature & 1.0 \\
Maximum response length & 8,192 \\
KL regularization & Disabled \\
\bottomrule
\end{tabular}
\caption{RL hyperparameters.}
\label{tab:rlvr-hyperparameters}
\end{minipage}
\end{table}
\FloatBarrier

\section{Additional Evaluation Results}

\subsection{Comparison with General-Purpose LLMs}

Table~\ref{tab:general-model-results} reports the complete results of general-purpose LLMs and \textsc{MathForm}-8B across all six benchmarks. DeepSeek-V4-Pro is evaluated under the \texttt{high} reasoning-effort setting.
\textsc{MathForm}-8B attains an average SC pass rate of 88.06\%, exceeding all
evaluated general-purpose models, including the strongest, Qwen3.7-Plus
(86.33\%); however, its
average CC pass rate (72.37\%) remains below Qwen3.7-Plus (83.38\%) and
DeepSeek-V4-Pro (76.54\%). These results highlight the capability--efficiency
trade-off of \textsc{MathForm}-8B: task-specific training yields strong
compilability in an 8B model, while frontier general-purpose models retain an
advantage in semantic fidelity.
\FloatBarrier
\begin{table}[htbp]
\centering
\small
\setlength{\tabcolsep}{2.2pt}
\resizebox{\textwidth}{!}{%
\begin{tabular}{@{}l*{14}{c}@{}}
\toprule
& \multicolumn{2}{c}{\textbf{AVG}}
& \multicolumn{2}{c}{\textbf{FormalMATH}}
& \multicolumn{2}{c}{\textbf{ProverBench}}
& \multicolumn{2}{c}{\textbf{CombiBench}}
& \multicolumn{2}{c}{\textbf{FATE-M}}
& \multicolumn{2}{c}{\textbf{FATE-H}}
& \multicolumn{2}{c}{\textbf{FATE-X}} \\
\cmidrule(lr){2-3}\cmidrule(lr){4-5}\cmidrule(lr){6-7}
\cmidrule(lr){8-9}\cmidrule(lr){10-11}\cmidrule(lr){12-13}
\cmidrule(lr){14-15}
\textbf{Model}
& \textbf{SC} & \textbf{CC}
& \textbf{SC} & \textbf{CC}
& \textbf{SC} & \textbf{CC}
& \textbf{SC} & \textbf{CC}
& \textbf{SC} & \textbf{CC}
& \textbf{SC} & \textbf{CC}
& \textbf{SC} & \textbf{CC} \\
\midrule
\multicolumn{15}{@{}l}{\textit{General-Purpose LLMs}} \\
\addlinespace[1pt]
DeepSeek-V4-Pro
& 78.34 & \underline{76.54}
& 97.88 & \underline{96.24}
& \underline{94.83} & 93.68
& 85.00 & \underline{81.00}
& 87.33 & 87.33
& 70.00 & \underline{68.00}
& \underline{35.00} & 33.00 \\
Qwen3.7-Plus
& \underline{86.33} & \textbf{83.38}
& \underline{99.29} & \textbf{98.59}
& \textbf{100.00} & \textbf{97.70}
& \underline{92.00} & \textbf{87.00}
& \underline{94.67} & \underline{92.00}
& \underline{78.00} & \textbf{77.00}
& \textbf{54.00} & \textbf{48.00} \\
Qwen3-235B-A22B-Thinking-2507
& 58.94 & 55.37
& 91.29 & 88.24
& 81.03 & 75.29
& 59.00 & 50.00
& 71.33 & 70.67
& 32.00 & 30.00
& 19.00 & 18.00 \\
Qwen3-32B
& 43.07 & 36.55
& 80.94 & 72.94
& 61.49 & 50.00
& 43.00 & 27.00
& 46.00 & 43.33
& 20.00 & 19.00
& 7.00 & 7.00 \\
DeepSeek-R1-0528-Qwen3-8B
& 47.28 & 37.65
& 72.24 & 61.18
& 52.30 & 40.23
& 36.00 & 15.00
& 36.00 & 30.00
& 4.00 & 4.00
& 4.00 & 1.00 \\
Qwen3-8B
& 27.25 & 17.06
& 61.88 & 42.12
& 37.93 & 27.59
& 29.00 & 7.00
& 26.67 & 22.67
& 4.00 & 3.00
& 4.00 & 0.00 \\
\midrule
\multicolumn{15}{@{}l}{\textit{Ours}} \\
\addlinespace[1pt]
\rowcolor{oursrow}
\textbf{\textsc{MathForm}-8B}
& \textbf{88.06} & 72.37
& \textbf{100.00} & 95.06
& \textbf{100.00} & \underline{94.83}
& \textbf{93.00} & 47.00
& \textbf{99.33} & \textbf{97.33}
& \textbf{82.00} & 63.00
& \textbf{54.00} & \underline{37.00} \\
\bottomrule
\end{tabular}%
}
\caption{Pass@8 SC and CC pass rates (\%) of general-purpose LLMs and \textsc{MathForm}-8B across six benchmarks. AVG is the equally weighted macro-average across all six benchmarks. The best result in each column is shown in \textbf{bold} and the second best is \underline{underlined}.}
\label{tab:general-model-results}
\end{table}
\FloatBarrier

\subsection{Human Evaluation}

For each problem in FATE-M and FATE-H, we randomly sample one candidate from the eight formalizations generated by each model and ask two human experts to assess whether it faithfully preserves the semantics of the source statement; disagreements are resolved through discussion between the two experts. Table~\ref{tab:human-evaluation} summarizes the results.
\FloatBarrier
\begin{table}[htbp]
\centering
\small
\setlength{\tabcolsep}{10pt}
\begin{tabular}{@{}lrrrr@{}}
\toprule
& \multicolumn{2}{c}{\textbf{FATE-M}}
& \multicolumn{2}{c}{\textbf{FATE-H}} \\
\cmidrule(lr){2-3}\cmidrule(lr){4-5}
\textbf{Model}
& \textbf{SC} & \textbf{CC}
& \textbf{SC} & \textbf{CC} \\
\midrule
StepFun-Formalizer-32B
& 42.67 & 36.00 & 16.00 & 12.00 \\
Goedel-Formalizer-V2-32B
& 75.33 & 68.00 & 38.00 & 27.00 \\
ReForm-32B
& 78.67 & 74.00 & 50.00 & 41.00 \\
\rowcolor{oursrow}
\textbf{\textsc{MathForm}-8B}
& \textbf{86.67} & \textbf{76.67}
& \textbf{54.00} & \textbf{42.00} \\
\bottomrule
\end{tabular}
\caption{Human-evaluation results (\%) on FATE-M and FATE-H, based on one randomly sampled candidate per problem. The best result in each column is shown in \textbf{bold}.}
\label{tab:human-evaluation}
\end{table}
\FloatBarrier

Human evaluation preserves the relative ordering of the models observed under
automated evaluation.
\textsc{MathForm}-8B attains the highest SC and CC pass rates on both FATE-M and
FATE-H, reaching 86.67\% and 76.67\% on FATE-M and 54.00\% and 42.00\% on
FATE-H.
These results indicate that the advantage of \textsc{MathForm}-8B on the
formalization of challenging statements is stable and also holds under human
assessment.

\section{Case Studies}
\label{sec:case-studies}

\subsection{Case Study of the Data Construction Pipeline}
\label{sec:case-pipeline}

We provide a detailed case study of the refinement pipeline on a group-theoretic
problem, recording the retrieval queries and results, the generated code, and
the feedback of both verification steps in each round.
\begin{itemize}
    \item \textbf{Round 1.} The retrieved results cover only the centralizer.
    Although the generated code compiles, the semantic consistency check finds
    that it formalizes ``$K$ is normal in $G$'' rather than the intended
    statement that ``$K$ is normal in $H$,'' thereby introducing a stronger
    assumption.
    \item \textbf{Round 2.} Guided by the semantic feedback, the system issues
    an additional query about subgroup normality. The retrieved result clarifies
    the scope of the corresponding Mathlib definition, enabling the model to
    revise the condition so that it correctly states normality relative to
    $H$. The revised code passes both checks.
\end{itemize}
This example shows that targeted retrieval in response to verification feedback
can supply the library knowledge needed to correct a semantic deviation and
guide the model toward a faithful formalization.

\begin{caseboxlong}{Case Study of the Data Construction Pipeline}
\casetag{Natural-language statement}
Let $G$ be a group and let $K \subseteq H$
be subgroups of $G$ with $K \triangleleft H$. If $H \triangleleft G$ and
$C_H(K) = 1$, prove that $H$ centralizes $C_G(K)$.

\tcbsubtitle{Round 1}

\casetag{Retrieval}
\begin{lstlisting}[style=prompt]
[1] Query: subgroup.centralizer definition
    Subgroup.centralizer      Mathlib/GroupTheory/Subgroup/Centralizer.lean:22
      def centralizer (s : Set G) : Subgroup G := { Submonoid.centralizer s with carrier := Set.centralizer s inv_mem' := Set.inv_mem_centralizer }
      The `centralizer` of `s` is the subgroup of `g : G` commuting with every `h : s`.
    Subgroup.center           Mathlib/GroupTheory/Subgroup/Center.lean:22
      def center : Subgroup G := { Submonoid.center G with carrier := Set.center G inv_mem' := Set.inv_mem_center }
      The center of a group `G` is the set of elements that commute with everything in `G`
\end{lstlisting}

\casetag{Generated formalization}
\begin{lstlisting}[language=lean4]
import Mathlib.GroupTheory.Subgroup.Centralizer
import Mathlib.GroupTheory.Subgroup.Basic

open Subgroup

variable {G : Type*} [Group G]

theorem H_centralizes_CG_K
  {K H : Subgroup G}
  (hK_le_H : K ≤ H)                     -- `K ⊆ H`
  (hK_normal : K.Normal)                -- `K ◁ H`
  (hH_normal : H.Normal)                -- `H ◁ G`
  (hC_eq_one : (Subgroup.centralizer (K : Set G) ⊓ H) = ⊥) :
    H ≤ Subgroup.centralizer (Subgroup.centralizer (K : Set G) : Set G) := by sorry
\end{lstlisting}

\caseverdict{\statusok{Compilation succeeded}\hspace{1.8em}\statusfail{Semantic consistency: incorrect}}

The Lean code incorrectly states \texttt{K.Normal} for the normality of $K$,
which implies $K$ is normal in $G$, whereas the NL specifies $K$ is normal in
$H$ ($K \triangleleft H$). All other aspects (centralizer conditions, $H$'s
normality in $G$, and the conclusion) are correctly formalized.

\tcbsubtitle{Round 2}

\casetag{Retrieval}
\begin{lstlisting}[style=prompt]
[2] Query: normal subgroup of a subgroup definition
    Subgroup.Normal           Mathlib/Algebra/Group/Subgroup/Defs.lean:591
      structure Normal : Prop where conj_mem : ∀ n, n ∈ H → ∀ g : G, g * n * g⁻¹ ∈ H
      A subgroup is normal if whenever `n ∈ H`, then `g * n * g⁻¹ ∈ H` for every `g : G`
    commutator                Mathlib/GroupTheory/Abelianization.lean:37
      def commutator : Subgroup G := ⁅(⊤ : Subgroup G), ⊤⁆
      The commutator subgroup of a group G is the normal subgroup generated by the commutators [p,q]=`p*q*p⁻¹*q⁻¹`.
\end{lstlisting}

\casetag{Generated formalization}
\begin{lstlisting}[language=lean4]
import Mathlib.GroupTheory.Subgroup.Centralizer
import Mathlib.GroupTheory.Subgroup.Basic

open Subgroup

variable {G : Type*} [Group G]

theorem H_centralizes_CG_K
  {K H : Subgroup G}
  (hK_le_H : K ≤ H)
  (hK_normal_in_H :
    ∀ g ∈ H, ∀ k ∈ K, g * k * g⁻¹ ∈ K)
  (hH_normal : H.Normal)
  (hC_eq_bot : (Subgroup.centralizer (K : Set G) ⊓ H) = ⊥) :
    H ≤ Subgroup.centralizer (Subgroup.centralizer (K : Set G) : Set G) := by sorry
\end{lstlisting}

\caseverdict[colback=okgreen!10]{\statusok{Compilation succeeded}\hspace{1.8em}\statusok{Semantic consistency: correct}}

The Lean 4 code accurately formalizes the NL statement. The hypotheses
correctly capture $K$ being a subgroup of $H$, normality of $K$ in $H$,
normality of $H$ in $G$, and the triviality of $C_H(K)$. All logical
dependencies and quantifiers align between the NL and the code. No semantic
mismatches detected.
\end{caseboxlong}

\subsection{Case Study of \textsc{MathForm}-8B}
\label{sec:case-mathform}

In this commutative-algebra case, the three 32B baselines all fail to compile
for the same reason: they refer to identifiers that do not exist in Mathlib,
such as \texttt{Frac} or \texttt{QuotField} for the quotient field.
\textsc{MathForm}-8B maps the quotient field to \texttt{FractionRing} and
expresses the existence of an isomorphism by wrapping a ring equivalence in
\texttt{Nonempty}, yielding a correct formalization.

\begin{caseboxlong}{Case Study of a Commutative-Algebra Problem}
\casetag{Natural-language statement}
Show that if $R$ is a unique factorization
domain whose quotient field is isomorphic to $\mathbb{R}$, then $R$ is
isomorphic to $\mathbb{R}$.

\tcbsubtitle{\textsc{MathForm}-8B}

\casetag{Model output}
\begin{lstlisting}[style=prompt]
<think>
**Mathematical concept understanding**

The natural-language statement talks about a *unique factorization domain* $R$ whose *quotient field* (i.e. fraction field) is isomorphic to the real numbers $\mathbb{R}$. The goal is to deduce that the ring $R$ itself is isomorphic to $\mathbb{R}$.

The objects that appear are:
- A commutative ring $R$ that is an integral domain and satisfies the unique factorisation property.
- The fraction field $\operatorname{FractionRing} R$, which is the field of fractions of $R$.
- A field isomorphism (ring equivalence) between $\operatorname{FractionRing} R$ and $\mathbb{R}$.
- A ring equivalence between $R$ and $\mathbb{R}$.

No ambiguity remains once we interpret "quotient field" as the standard fraction field construction and "isomorphic to $\mathbb{R}$" as a field isomorphism. The implicit conventions are:
- $\mathbb{R}$ is the real field, denoted `ℝ` in Lean.
- The isomorphism between fields is a *ring equivalence* (type `≃+*`).

**Logical structure analysis**

The statement is a universal implication:
$\displaystyle \forall R\;(\text{UFD }R)\;(\text{FractionRing }R \cong \mathbb{R})\;\Longrightarrow\;R \cong \mathbb{R}.$
Formally we have:
- Hypothesis `hK : (FractionRing R) ≃+* ℝ`.
- Conclusion `Nonempty (R ≃+* ℝ)`.

Thus the theorem is a single implication with a single hypothesis and a single conclusion.

**Formalization strategy**

*Types for the objects*
- `R` must be a type equipped with `[CommRing R]` and `[IsDomain R]`.
- The unique factorisation property is expressed by the typeclass `[UniqueFactorizationMonoid R]`.
- The fraction field is `FractionRing R`.
- A field isomorphism is a `RingEquiv`, written `≃+*`.
- To express existence of an isomorphism we wrap the `RingEquiv` in `Nonempty`.

*Imports*
The following modules provide the needed definitions and instances:
- `Mathlib.RingTheory.FractionRing` for `FractionRing`.
- `Mathlib.Algebra.Algebra.Basic` for basic algebraic structures.
- `Mathlib.Data.Real.Basic` for the real numbers `ℝ`.
- `Mathlib.RingTheory.UniqueFactorizationDomain` for the `UniqueFactorizationMonoid` typeclass.
- `Mathlib.RingTheory.Localization.FractionRing` for the localization construction.

*Theorem shape*
We formulate a theorem named `my_favorite_theorem`. Its arguments are:
- An implicit type `R`.
- The three typeclass arguments listed above.
- The hypothesis `hK : (FractionRing R) ≃+* ℝ`.

The conclusion is `Nonempty (R ≃+* ℝ)`. The proof placeholder is `by sorry`.

*Auxiliary definitions*
No auxiliary lemmas are needed for the statement itself; the theorem is a pure existence claim.

**Syntax translation**

1. **Variable declaration**
   ```lean
   {R : Type*}
   ```
   `R` is an implicit type argument.

2. **Typeclass arguments**
   ```lean
   [CommRing R] [IsDomain R] [UniqueFactorizationMonoid R]
   ```
   These are placed after `R` and before the hypothesis.

3. **Hypothesis**
   ```lean
   (hK : (FractionRing R) ≃+* ℝ)
   ```
   `FractionRing R` is the fraction field; `≃+*` is the notation for `RingEquiv`.

4. **Conclusion**
   ```lean
   Nonempty (R ≃+* ℝ)
   ```
   `Nonempty` expresses that at least one `RingEquiv` exists.

5. **Theorem header**
   ```lean
   theorem my_favorite_theorem
   ```
   The name follows the usual lower-snake-case convention.

6. **Proof placeholder**
   ```lean
   : Nonempty (R ≃+* ℝ) := by sorry
   ```
   The `by sorry` term tells Lean that the proof is omitted.

**Complete Lean 4 code**

```lean
import Mathlib.RingTheory.FractionRing
import Mathlib.Algebra.Algebra.Basic
import Mathlib.Data.Real.Basic
import Mathlib.RingTheory.UniqueFactorizationDomain
import Mathlib.RingTheory.Localization.FractionRing

open scoped Classical

theorem my_favorite_theorem
  {R : Type*} [CommRing R] [IsDomain R] [UniqueFactorizationMonoid R]
  (hK : (FractionRing R) ≃+* ℝ) :
  Nonempty (R ≃+* ℝ) := by sorry
```

This code follows directly from the reasoning above: the imports bring in the necessary algebraic constructions, the typeclass assumptions capture the unique factorisation domain hypothesis, the hypothesis `hK` encodes the isomorphism of fraction fields with the reals, and the conclusion asserts the existence of a ring isomorphism between `R` and `ℝ`. No further modifications are required.
</think>
\end{lstlisting}

\begin{lstlisting}[language=lean4]
import Mathlib.RingTheory.FractionRing
import Mathlib.Algebra.Algebra.Basic
import Mathlib.Data.Real.Basic
import Mathlib.RingTheory.UniqueFactorizationDomain
import Mathlib.RingTheory.Localization.FractionRing

open scoped Classical

theorem my_favorite_theorem
  {R : Type*} [CommRing R] [IsDomain R] [UniqueFactorizationMonoid R]
  (hK : (FractionRing R) ≃+* ℝ) :
  Nonempty (R ≃+* ℝ) := by sorry
\end{lstlisting}

\caseverdict[colback=okgreen!10]{\statusok{Compilation succeeded}\hspace{1.8em}\statusok{Semantic consistency: correct}}

\tcbsubtitle{Goedel-Formalizer-V2-32B}
\begin{lstlisting}[language=lean4]
import Mathlib
import Aesop

set_option maxHeartbeats 0

open BigOperators Real Nat Topology Rat

theorem my_favorite_theorem (R : Type*) [CommRing R] [IsDomain R] [UniqueFactorizationMonoid R]
  (h_quotient : Nonempty (Frac R ≃+* ℝ)) :
  Nonempty (R ≃+* ℝ) := by sorry
\end{lstlisting}
\caseverdict[colback=failred!10]{\statusfail{Compilation failed}}
\texttt{Frac} is not applicable as a function.

\tcbsubtitle{StepFun-Formalizer-32B}
\begin{lstlisting}[language=lean4]
import Mathlib

theorem my_favorite_theorem {R : Type*} [CommRing R] [IsUFD R]
    (h : ∃ (f : QuotField R ≃ ℝ)) : ∃ (g : R ≃ ℝ) := by
  sorry
\end{lstlisting}
\caseverdict[colback=failred!10]{\statusfail{Compilation failed}}
The generated theorem contains a syntax error in the hypothesis.

\tcbsubtitle{ReForm-32B}
\begin{lstlisting}[language=lean4]
import Mathlib
import Aesop

set_option maxHeartbeats 0

open BigOperators Real Nat Topology Rat

theorem my_favorite_theorem (R : Type*) [CommRing R] [IsDomain R] [UniqueFactorizationMonoid R]
  (h : Nonempty (Frac R ≃+* ℝ)) :
  Nonempty (R ≃+* ℝ) := by sorry
\end{lstlisting}
\caseverdict[colback=failred!10]{\statusfail{Compilation failed}}
\texttt{Frac} is not applicable as a function.
\end{caseboxlong}

\section{Prompts}
\label{sec:prompts}

\subsection{Data Construction Prompts}
The two prompts implement the formalization generator and the retrieval planner
of the pipeline described in the Method section. During refinement rounds, both
prompts additionally carry the compiler diagnostics and semantic-consistency
feedback from the previous attempt.

\begin{promptbox}{Formalization Generator Prompt}
You are an expert in Lean 4 theorem proving and the Mathlib mathematical library.
Your task is to formalize the given mathematical statement into correct Lean 4 code using Mathlib.

FORMALIZATION REQUIREMENTS:
1. Use proper Lean 4 syntax and Mathlib conventions
2. Include ALL necessary headers
3. Define appropriate variables and assumptions
4. The theorem statement must be mathematically correct and equivalent to the original
5. This task is only about automatic formalization. Do NOT output any proof steps, tactics, or reasoning.
6. Retrieved information may be helpful; identify what is truly relevant and use it as needed.
7. Only generate the translation. Do not try to solve or prove the problem.

RETRIEVED MATHLIB INFORMATION:
{retrieval_context}

PREVIOUS COMPILATION FAILURE - FIX REQUIRED:
Failed code:
{last_bad_code}
Compiler error:
{compile_error}
Please analyze the error and provide corrected code.

SEMANTIC CONSISTENCY ISSUE - FIX REQUIRED:
Previous code:
{last_bad_code}
Semantic feedback:
{semantic_feedback}
Please revise the formalization to match the mathematical meaning.

STATEMENT TO FORMALIZE:
{statement}
\end{promptbox}

\begin{promptbox}{Retrieval Planner Prompt}
You are a Lean 4/Mathlib expert. I need to formalize the mathematical statement below into Lean 4 code.
To ensure accurate formalization, I will search the Lean mathematical library (Mathlib) for relevant definitions and theorems.

First review the existing queries and retrieved results to decide whether any new search queries are truly necessary. Only generate new queries if they are essential to resolve ambiguity or missing definitions. If more info is needed, generate 1-3 new queries; otherwise output nothing.

Each query should be:
- Concise and specific
- Focus on mathematical concepts, definitions, or theorems

Do not repeat any previous queries.
Each query must be highly necessary and directly helpful for formalization; avoid broad or speculative queries.
If no additional queries are needed, output an EMPTY code block only.

Output format: output ONLY a single triple-backtick block. One query per line, no numbering or bullets. No extra text, no explanations.

Examples:
```
group definition
ring homomorphism
topological space compactness
natural number induction
vector space dimension
```

Empty output example (no additional queries needed):
```
```

Previous queries:
{previous_queries}

Retrieved results so far:
{retrieval_context}

Previous verification feedback:
- Compilation error: {compile_error}
- Semantic feedback: {semantic_feedback}

Statement to formalize:
{statement}
\end{promptbox}

\subsection{Semantic Consistency Check Prompt}
The semantic consistency check compares a natural-language statement with its
Lean~4 formalization and returns a binary judgment. The same prompt is used for
validation during data construction, for the semantic term of the reward, and
for evaluation.

\begin{promptbox}{Semantic Consistency Check Prompt}
# Lean 4 Formalization Semantic Consistency Check

**Role:**
Act as an expert in Lean 4 formal verification and mathematical logic. Your task is to perform a rigorous semantic consistency review, comparing a Natural Language (NL) mathematical statement against its corresponding Lean 4 formalization.

**Objective:**
Determine if the provided Lean 4 code is a faithful, completely accurate, and logically equivalent translation of the Natural Language statement. Only focus on the formalization itself, do not discuss any proof process.

**Analysis Steps:**
Before generating the output, analyze the input pairs based on the following criteria:

1. **Deconstruction of the Natural Language Statement:**
   * Identify all key mathematical objects, definitions, and properties.
   * Map out the logical flow (antecedents, consequents, quantifiers).
   * Identify the implicit domain of discourse.

2. **Analysis of the Lean 4 Code Structure:**
   * Verify type hierarchy compliance (classes vs. structures).
   * Check variable declarations and hypothesis scope.
   * Ensure standard library usage matches the mathematical intent.

3. **Semantic Mapping and Gap Analysis:**
   * **Bi-directional Fidelity:** Ensure every constraint in the NL maps to the code, and the code adds no unintended constraints.
   * **Quantifier Precision:** Rigorously check the order and dependency of `for all` vs `there exists`.
   * **Condition Strength:** Ensure predicates are neither strictly stronger nor strictly weaker than required.

**Output Format Requirements:**
Your output must be **exactly** two XML tags in this order, with no other text before, between, or after them. Do not use code fences or markdown.

1. <comments></comments>
   * Provide your detailed evaluation and reasoning.
   * If inconsistent, describe the specific semantic mismatch.
2. <result></result>
   * Output exactly one word: `correct` or `incorrect`.

**Strict Output Template:**
<comments>...</comments>
<result>...</result>

---

**Input Natural Language Statement:**
{mathematical_statement}

**Input Lean 4 Code:**
{autoformalization_placeholder}
\end{promptbox}

\subsection{Trajectory Reconstruction Prompt}
Trajectory reconstruction takes a verified natural-language statement together
with its Lean~4 formalization as input and synthesizes a formalization
trajectory retrospectively. The prompt organizes this process into four stages,
namely concept understanding, logical structure analysis, formalization
strategy, and syntactic translation, and restricts its content to formalization
strategy rather than proof strategy.

\begin{promptbox}{Trajectory Reconstruction Prompt}
Please read the following natural language math problem and its Lean 4 formalization below. Then reconstruct the thinking process a mathematician would go through when formalizing this problem into Lean 4. Your response should mimic the actual reasoning flow, including:

**Key Requirements for Authentic CoT:**

1. **Mathematical Concept Understanding**: What mathematical concepts and entities appear in this problem? What are the key mathematical objects involved (numbers, functions, sets, structures, etc.)? Are there any ambiguities in the natural language description that need clarification? What implicit assumptions or conventions might be present? Consider domain constraints, well-definedness conditions, and mathematical context.

2. **Logical Structure Analysis**: Analyze the logical form of the statement. Is it a conditional (implication), biconditional (equivalence), universal quantification, existential statement, or a combination? Identify the hypotheses and conclusion clearly. What is the dependency structure between different parts of the statement?

3. **Formalization Strategy**: 
   - Determine appropriate types for each mathematical object. Should variables be ℝ, ℕ, ℤ, ℚ, or more complex types? Provide a concise justification based on the mathematical operations and constraints.
   - Outline the overall formalization approach: What theorem structure is needed? What key imports are required? Are auxiliary definitions or lemmas necessary? Should this be formulated as a theorem, lemma, or definition?
   - Briefly plan how to represent complex mathematical structures (e.g., matrices, sequences, sets) in Lean's type system.

4. **Syntax Translation**: Now translate each component step by step:
   - How should each mathematical object be declared? Consider implicit vs explicit arguments, type annotations needed.
   - How to express each hypothesis in Lean syntax? Walk through each condition, explaining notation choices (e.g., how to write matrix entries, how to express distinctness, how to state equations).
   - How to formalize mathematical operations? Consider operator precedence, parenthesization, and Lean-specific syntax requirements.
   - How to structure the theorem statement? Discuss the order of hypotheses, naming conventions, and how the conclusion is formulated.
   - Address any subtle translation issues: type coercions, implicit arguments, namespace qualifications, notation systems.
   - Present the complete Lean 4 code that results from the above reasoning process.

**Critical Stylistic Guidelines:**

- Do NOT question or doubt the accuracy of the provided Lean 4 code, nor raise concerns about it, don't make any modifications to the provided code
- Avoid meta-statements like "the provided code" or "the given code"
- Focus entirely on **formalization strategy**, NOT proof strategy
- Show genuine reasoning with considerations of alternatives and justifications for choices made
- Be specific about Lean syntax decisions rather than giving high-level descriptions
- Please try to describe the content using plain text as much as possible, and avoid using the table format of markdown

### Natural Language Problem
{statement}

### Lean 4 Code
```Lean4
{lean_code}
```
\end{promptbox}

\subsection{Evaluation Inference Prompts}
Each model is evaluated with its officially recommended prompt template. For
general-purpose LLMs, we additionally include an explicit instruction not to
produce a proof, which prevents instruction drift during inference.

\begin{promptbox}{\textsc{MathForm}-8B Inference Prompt (Ours)}
Please convert the following informal math problem to a formal one in Lean 4 with a header. Use the following theorem names: my_favorite_theorem.

{informal_problem}
\end{promptbox}

\begin{promptbox}{General-Purpose LLM Inference Prompt}
Please convert the following informal math problem to a formal one in Lean 4 with a header. Do not provide the proof, end with `by sorry`. Use the following theorem names: my_favorite_theorem.

{informal_problem}
\end{promptbox}

\begin{promptbox}{Kimina-Autoformalizer Inference Prompt}
You are an expert in mathematics and Lean 4. Please autoformalize the following problem in Lean 4 with a header. Use the following theorem names: my_favorite_theorem.

{informal_problem}
\end{promptbox}

\begin{promptbox}{Mathesis Inference Prompt}
[Question]:
{informal_problem}

You are an expert in formal mathematics. Your task is to convert the above [question] to lean 4
theorems by completing the following lean 4 code:

```lean4
import Mathlib
import Aesop
set-option maxHeartbeats 0
set-option pp.numericTypes true
set-option pp.coercions true
set-option pp.letVarTypes true
set-option pp.structureInstanceTypes true
set-option pp.instanceTypes true
set-option pp.mvars.withType true
set-option pp.coercions true
set-option pp.funBinderTypes true
set-option pp.piBinderTypes true
open BigOperators Real Nat Topology Rat

/-
{informal_problem}
-/
```
\end{promptbox}

\begin{promptbox}{StepFun-Formalizer Inference Prompt}
Please autoformalize the following problem in Lean 4 with a header. Use the following theorem names: my_favorite_theorem.

{informal_problem}

Your code should start with:
```Lean4
import Mathlib
```
\end{promptbox}

\begin{promptbox}{Goedel-Formalizer-V2 Inference Prompt}
Please autoformalize the following natural language problem statement in Lean 4. Use the following theorem name: my_favorite_theorem
The natural language statement is: 
{informal_problem}Think before you provide the lean statement.
\end{promptbox}

\begin{promptbox}{ReForm Inference Prompt}
Think step by step to translate the mathematical problem in natural language to Lean 4, and verify the consistency.
{informal_problem}
\end{promptbox}

\end{document}